\documentclass[10pt,twocolumn,letterpaper]{article}

\usepackage{cvpr}              

\usepackage{graphicx}
\usepackage{amsmath}
\usepackage{amssymb}
\usepackage{makecell}
\usepackage{colortbl}
\usepackage[table]{xcolor}
\usepackage{booktabs}
\usepackage{multirow}
\usepackage{enumitem}
\setlist[itemize]{leftmargin=*}
\setenumerate[1]{itemsep=0pt,partopsep=0pt,parsep=\parskip,topsep=5pt}
\setitemize[1]{itemsep=0pt,partopsep=0pt,parsep=\parskip,topsep=5pt}
\setdescription{itemsep=0pt,partopsep=0pt,parsep=\parskip,topsep=5pt}

\usepackage{bbding}
 
\usepackage[accsupp]{axessibility}

\usepackage[pagebackref,breaklinks,colorlinks]{hyperref}

\usepackage[capitalize]{cleveref}
\crefname{section}{Sec.}{Secs.}
\Crefname{section}{Section}{Sections}
\Crefname{table}{Table}{Tables}
\crefname{table}{Tab.}{Tabs.}
\usepackage{flushend}
\usepackage{balance}

\def\cvprPaperID{*****} 
\def\confName{CVPR}
\def\confYear{2023}
\def\cvprPaperID{497}

\begin{document}

\title{V2X-Seq: A Large-Scale Sequential Dataset for  \\
Vehicle-Infrastructure Cooperative Perception and Forecasting}
\author{Haibao Yu$^{1,2}$, Wenxian Yang$^{1}$, Hongzhi Ruan$^{1,5}$, Zhenwei Yang$^{1,6}$, Yingjuan Tang$^{1,7}$, Xu Gao$^{3}$, Xin Hao$^{3}$, 
\and Yifeng Shi$^{3}$, Yifeng Pan$^{3}$, Ning Sun$^{4}$, Juan Song$^{4}$, Jirui Yuan$^{1}$, Ping Luo$^{2}$, Zaiqing Nie$^{1}$~\thanks{Corresponding author. Work done while at AIR. For any questions or discussions, please email dair@air.tsinghua.edu.cn.} \\
$^{1}$Institute for AI Industry Research (AIR), Tsinghua University $^{2}$The University of Hong Kong \\
$^{3}$Baidu Inc. $^{4}$Beijing Connected and Autonomous Vehicles Technology Co., Ltd \\
$^{5}$University of Chinese Academy of Science \\
$^{6}$University of Science and Technology Beijing 
$^{7}$Beijing Institute of Technology \\
}

\maketitle

\begin{abstract}
Utilizing infrastructure and vehicle-side information to track and forecast the behaviors of surrounding traffic participants can significantly improve decision-making and safety in autonomous driving. However, the lack of real-world sequential datasets limits research in this area. To address this issue, we introduce V2X-Seq, the first large-scale sequential V2X dataset, which includes data frames, trajectories, vector maps, and traffic lights captured from natural scenery. V2X-Seq comprises two parts: the sequential perception dataset, which includes more than 15,000 frames captured from 95 scenarios, and the trajectory forecasting dataset, which contains about 80,000 infrastructure-view scenarios, 80,000 vehicle-view scenarios, and 50,000 cooperative-view scenarios captured from 28 intersections' areas, covering 672 hours of data.
Based on V2X-Seq, we introduce three new tasks for vehicle-infrastructure cooperative (VIC) autonomous driving: VIC3D Tracking, Online-VIC Forecasting, and Offline-VIC Forecasting. We also provide benchmarks for the introduced tasks.
Find data, code, and more up-to-date information
at \href{https://github.com/AIR-THU/DAIR-V2X-Seq}{https://github.com/AIR-THU/DAIR-V2X-Seq}.
\end{abstract}

\section{Introduction}
Although single-vehicle autonomous driving has made significant advancements in recent years, it still faces significant safety challenges due to its limited perceptual field and inability to accurately forecast the behaviors of traffic participants. 
These challenges hinder autonomous vehicles from making well-informed decisions and driving safer. 
A promising solution to address these challenges is to leverage infrastructure information via Vehicle-to-Everything (V2X) communication, which has been shown to significantly expand perception range and enhance autonomous driving safety~\cite{yu2022dair,arnold2020cooperative}. 
However, current research primarily focuses on utilizing infrastructure data to improve the perception ability of autonomous driving, particularly in the context of frame-by-frame 3D detection. 
To enable well-informed decision-making for autonomous vehicles, it is critical to also incorporate infrastructure data to track and predict the behavior of surrounding traffic participants.

\begin{figure}[t]
	\centering
	\includegraphics[width=0.48\textwidth]{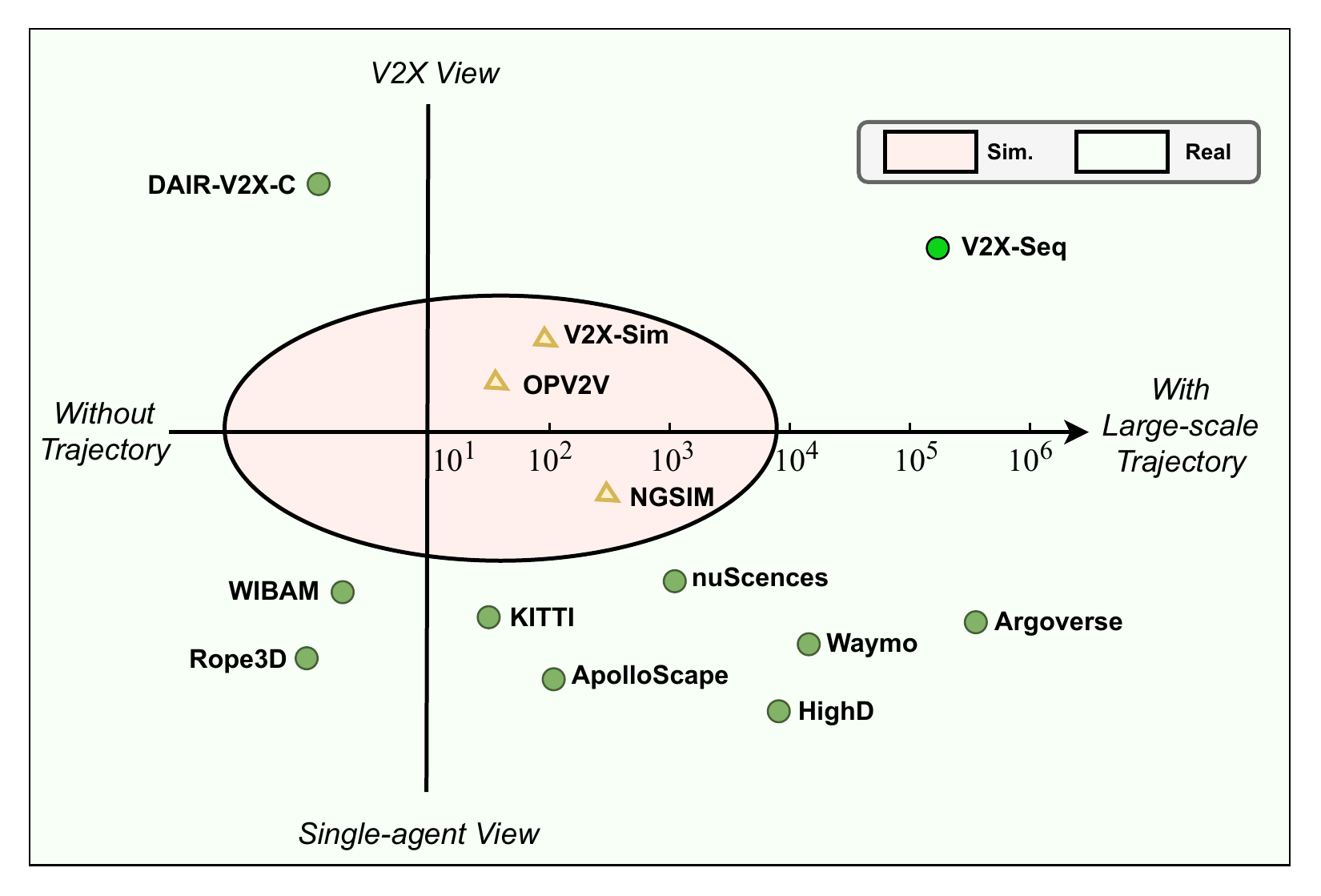}
	\caption{Autonomous driving datasets. V2X-Seq is the first large-scale, real-world, and sequential V2X dataset. The green circle denotes the real-world dataset, and the pink triangle denotes the simulated dataset. The abscissa represents the number of sequences.}
	\label{fig:positive_effect_vic3d}
\end{figure}
\begin{table*}[ht]
\caption{
Comparison with the public autonomous driving dataset. '-' denotes that the information is not provided. 'Real/Sim.' indicates whether the data was collected from the real world or a simulator. V2X view includes multi-vehicle cooperative view and vehicle-infrastructure cooperative view. V2X-Seq is the first large-scale sequential V2X dataset and focuses on vehicle-infrastructure cooperative view. All data elements, including the traffic light signals, are captured and generated from the real world.
}
\renewcommand\arraystretch{1.15}
\label{tab: dataset comparison}
\resizebox{\textwidth}{!}{%
\begin{tabular}{lccccccccccc}
\hline
\hline
\multirow{2}*{\textbf{Dataset}}
 & \multirow{2}*{\textbf{Year}} & \multirow{2}*{\textbf{Real/Sim.}} & \multirow{2}*{\textbf{View}} &  \textbf{With }&\textbf{With } & \textbf{With } & \textbf{With } & \textbf{Tracked }& \textbf{Total } &  \multirow{2}*{\textbf{Scenes}}\\
 &  &  & &  \textbf{ Trajectory}&\textbf{ 3D Boxes} & \textbf{ Maps} & \textbf{ Traffic Light} & \textbf{ Objects/Scene}& \textbf{Time (hour)} &\\
\hline
KITTI\cite{Geiger2012KITTI}  & 2012 & Real   & Single-vehicle                       & \Checkmark & \Checkmark &  \XSolid &  \XSolid  &43.67&1.5&50\\
\hline
nuScenes \cite{cae2020nus}     & 2019 & Real      & Single-vehicle                        & \Checkmark & \Checkmark & \Checkmark& \XSolid &75.75&5.5&1,000 \\
\hline
Waymo Motion~\cite{sun2020scalability, ettinger2021large}    & 2021 & Real      & Single-vehicle                       & \Checkmark& \Checkmark & \Checkmark  & \Checkmark &  - & 574 & 103,354 \\
\hline
Argoverse \cite{chang2019argoverse}   & 2019 & Real & Single-vehicle     & \Checkmark& \XSolid   & \Checkmark& \XSolid  &50.03&320&324,557\\
\hline
ApolloScape \cite{huang2019apolloscape,ma2019trafficpredict}  & 2019 & Real & Single-vehicle     & \Checkmark & \XSolid & \XSolid & \XSolid  &50.6&2.5&103\\
\hline
HighD \cite{krajewski2018highd} & 2018 & Real      & Drone                    & \Checkmark& \XSolid   & \Checkmark& \XSolid &-&16.5&5,940\\
\hline
WIBAM\cite{howe2021weakly} &2021&Real&Infrastructure& \XSolid  &\XSolid& \XSolid  &\XSolid& 0 &0.25& 0 \\
\hline

NGSIM  \cite{us2016next}   & 2016 & Sim.  & Infrastructure                      & \Checkmark & \XSolid & \XSolid & \XSolid&-&1.5&540 \\
\hline
V2X-Sim 2.0 \cite{li2022v2x} & 2022 & Sim.     & V2X                     & \Checkmark& \Checkmark & \XSolid  &\XSolid &-&0.3&100 \\
\hline
OPV2V \cite{xu2022opv2v}  & 2021 & Sim.      & V2X                      & \Checkmark& \Checkmark & \XSolid  &\XSolid & 26.5&0.2&73 \\
\hline
Cooper(inf)\cite{arnold2020cooperative} &2019&Sim.&V2X&  \Checkmark& \Checkmark & \XSolid  &\XSolid&30 &-& \textless100\\
\hline
DAIR-V2X-C \cite{yu2022dair}  & 2021 & Real      & V2X                        & \XSolid  & \Checkmark & \Checkmark & \XSolid & 0 & 0.5 & 100\\
\hline
\hline
\textbf{V2X-Seq/Perception}  & 2023 & Real      & V2X              &\Checkmark & \Checkmark  & \Checkmark  & \XSolid & 110 & 0..43 & 95 \\
\hline
\textbf{V2X-Seq/Forecasting}     & 2023 & Real  & V2X                 & \Checkmark& \Checkmark   & \Checkmark& \Checkmark & 101 & 583 &210,000 \\

\hline
\hline
\end{tabular}%
}
\end{table*}

To accelerate the research on cooperative sequential perception and forecasting, we release a large-scale sequential V2X dataset, V2X-Seq. All elements of this dataset were captured and generated from real-world scenarios. Compared with DAIR-V2X~\cite{yu2022dair}, which focuses on 3D object detection tasks, V2X-Seq is specifically designed for tracking and trajectory forecasting tasks.
The V2X-Seq dataset is divided into two parts: the sequential perception dataset and the trajectory forecasting dataset. 
The sequential perception dataset comprises 15,000 frames captured from 95 scenarios, which include infrastructure images, infrastructure point clouds, vehicle-side images, vehicle-side point clouds, 3D detection/tracking annotations, and vector maps. 
The trajectory forecasting dataset comprises 210,000 scenarios, including 50,000 cooperative-view scenarios, that were mined from 672 hours of data collected from 28 intersection areas.
To our knowledge, V2X-Seq is the first sequential V2X dataset that includes such a large-scale scenarios, making it an ideal resource for developing and testing cooperative perception and forecasting algorithms.

Based on the V2X-Seq dataset, we introduce three novel tasks for vehicle-infrastructure cooperative perception and forecasting. The first task is VIC3D Tracking, which aims to cooperatively locate, identify, and track 3D objects using sequential sensor inputs from both the vehicle and infrastructure. The second task is Online-VIC trajectory forecasting, which focuses on accurately predicting future behavior of target agents by utilizing past infrastructure trajectories, ego-vehicle trajectories, real-time traffic lights, and vector maps. The third task is Offline-VIC trajectory forecasting, which involves extracting relevant knowledge from previously collected infrastructure data to facilitate vehicle-side forecasting. These proposed tasks are accompanied by rich benchmarks. Additionally, we propose an intermediate-level framework, FF-Tracking, to effectively solve the VIC3D Tracking task.

The main contributions are organized as follows:
\begin{itemize}
    \item We release the V2X-Seq dataset, which constitutes the first large-scale sequential V2X dataset. All data are captured and generated from the real world.
    \item Based on the V2X-Seq dataset, we introduce three tasks for the vehicle-infrastructure cooperative autonomous driving community. To enable a fair evaluation of these tasks, we have carefully designed a set of benchmarks.
    \item We propose a middle fusion method, named FF-Tracking, for solving VIC3D Tracking and our proposed method can efficiently overcome the latency challenge.
\end{itemize}

\section{Related Work}
\paragraph{Autonomous Driving Datasets.}
Public datasets have greatly facilitated the development of autonomous driving. Kitti \cite{Geiger2012KITTI} is the pioneering dataset for autonomous driving. nuScenes \cite{cae2020nus}, Waymo Open \cite{sun2020scalability,ettinger2021large}, ApolloScape \cite{huang2019apolloscape}, and ONCE \cite{mao2021one} are large-scale and real-world datasets that support 3D object detection, tracking and prediction tasks. Argoverse \cite{chang2019argoverse}, Argoverse 2.0 \cite{wilson2021argoverse}, Lyft \cite{houston2021one}, and nuPlan \cite{caesar2021nuplan} release large-scale trajectories generated from the raw sensor data to support motion prediction and planning tasks. These datasets are all captured with single-vehicle sensors.
Repo3D \cite{ye2022rope3d}, WIBAM \cite{howe2021weakly}, and A9-Dataset \cite{cress2022a9} release the infrastructure-only 3D detection dataset. HighD \cite{krajewski2018highd} and NGSIM \cite{us2016next} release the drone or infrastructure-only trajectories dataset. OpenV2V \cite{xu2022opv2v}, V2X-Sim 2.0 \cite{li2022v2x}, and Cooper(inf) \cite{arnold2020cooperative} release small-scale sequential and simulated datasets for multi-vehicle cooperative perception.
DAIR-V2X-C \cite{yu2022dair} is the first real-world V2X dataset that supports VIC3D object detection; however, it does not provide the trajectory information. Compared with these existing public autonomous driving datasets, our V2X-Seq is the first large-scale sequential V2X dataset. All data are captured and generated from the real world. The dataset also includes vector maps and real-time traffic light signal data. It will be suitable for studying the Vehicle-Infrastructure Cooperative sequential perception and trajectory forecasting tasks.

\paragraph{Cooperative Autonomous Driving.}
Utilizing data from the road environment to enhance the safety of autonomous driving has attracted significant research attention in recent years. Some research works have focused on multi-vehicle cooperative perception, where lightweight feature-level data is transmitted and shared for improved perception of other vehicles \cite{Wang2020V2VNetVC,li2021learning,chen2019cooper}. To address communication delays in multi-vehicle 3D object detection, \cite{lei2022latency} proposes a time-compensation module for latency. On the other hand, some works have explored the use of infrastructure data to improve autonomous driving. For instance, \cite{yu2022dair} formalizes the vehicle-infrastructure cooperative 3D object detection task and highlights the latency challenges in cooperative perception.  \cite{yu2023vehicle} further proposed to use feature flow prediction to overcome the uncertain latency. Other works such as \cite{liu2020who2com,liu2020when2com,arnold2020cooperative,howe2021weakly} also consider transmitting feature-level data from infrastructure to the vehicle side. To empower only sharing sparse yet perceptually critical information, \cite{hu2022where2comm} utilizes a spatial confidence map. Moreover, \cite{li2022v2x} applies the Transformer \cite{vaswani2017attention} to fuse the features. Works such as \cite{Qiu2022AutoCastSI,cui2022coopernaut,valiente2019controlling} integrate the infrastructure data for control in autonomous driving.
However, most current works on cooperative autonomous driving focus on perceptual completion, overlooking the importance of temporal perception and forecasting. In this paper, we contribute to this field by releasing the V2X-Seq dataset, which is suitable for exploring sequential perception and forecasting tasks in vehicle-infrastructure cooperative settings.

\section{V2X-Seq Dataset}
To enable the exploration of the role of infrastructure in sequential perception and trajectory forecasting, we introduce the V2X-Seq dataset. This large-scale, real-world dataset contains sequential vehicle-to-everything (V2X) data. The sequential perception component of the dataset is presented in Section~\ref{sec:tem-pd}, while the trajectory forecasting component is detailed in Section~\ref{sec:tra-pd}. Additionally, we provide an overview of the vector maps and traffic lights used in the dataset in Section~\ref{sec:hd-map}.

\subsection{The Sequential Perception Dataset.}\label{sec:tem-pd}
3D tracking is a critical component in autonomous driving, as it provides sequential perception information that facilitates 3D detection and prediction. To enable exploration of the role of infrastructure in 3D tracking, we release the Sequential Perception Dataset (SPD). 
The SPD builds on the DAIR-V2X-C 3D detection dataset \cite{yu2022dair} and consists of more than 15,000 frames captured from 95 representative scenes with 10$\sim$20s duration sequences, comprising both vehicle sequential frames (images and point clouds) and infrastructure sequential frames (images and point clouds) sampled at 10 Hz. We provide 3D tracking annotations for each object of interest in each sequence, with unique tracking IDs shared by the same objects in each sequence, even if they are fully occluded in some frames. Additionally, for each scene, we provide an extra local vector map.

\paragraph{Data Collection and Annotation.} 
The SPD builds on the DAIR-V2X-C~\cite{yu2022dair}. We select 95 representative scenes from this dataset, where an autonomous driving vehicle drives through intersections equipped with sensors. SPD provides high-quality 3D annotations for ten object classes in every image and point cloud frame, including category attributes, occlusion state, truncated state, and a 7-dimensional cuboid modelled as x, y, z, width, length, height, and yaw angle. The object categories include various vehicles, pedestrians, and cyclists.
Building upon the DAIR-V2X-C dataset, our annotators assigned a unique tracking ID to each annotated object, except for static traffic cone objects. The same object in one sequence is assigned a unique tracking ID, even when it is completely occluded in some frames. Moreover, we provide cooperative tracking annotations for the cooperative-view sequences based on spatial and temporal matching.
Specifically, for each frame in each ego-vehicle sequence, we generate an infrastructure frame with the same timestamp as the corresponding ego-vehicle frame. This frame contains the 3D boxes interpolated and estimated from the infrastructure trajectories. Next, we convert these 3D boxes into an ego-vehicle coordinate system and match and fuse the two-side 3D boxes based on the Euclidean distance measurement and the Hungarian method~\cite{Kuhn2010TheHM}. To account for possible calibration and interpolation precision errors that may cause spatial matching errors, we compute the similarity of the two-side trajectories corresponding to the two matched 3D boxes. We filter out the matching with low scores and manually refine them to obtain accurate cooperative tracking annotations.

\begin{figure}[t]
	\centering
	\includegraphics[width=0.45\textwidth]{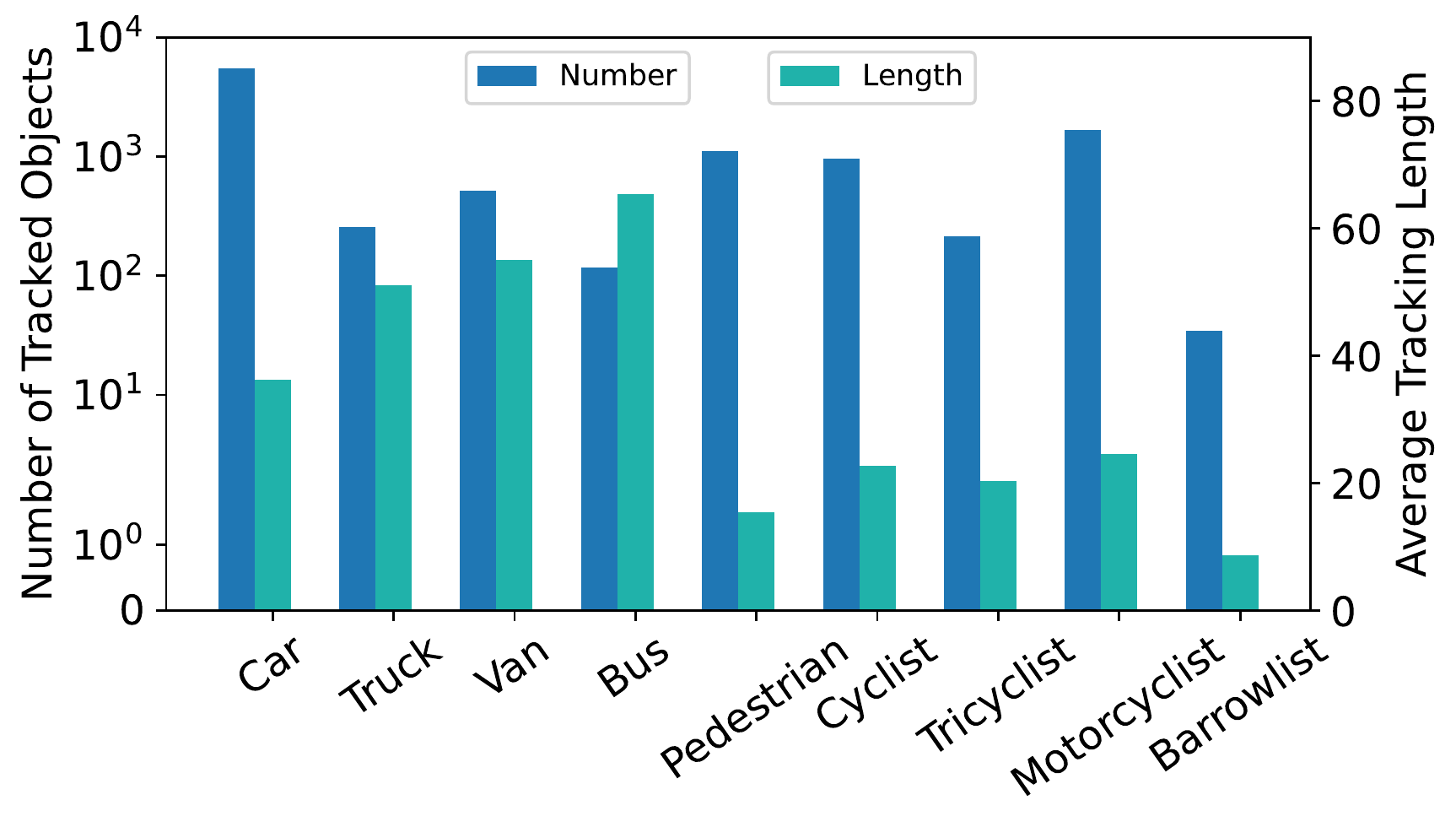}
    \vspace{-15pt}
	\caption{Total number and average tracking length of 3D tracked objects per category for the sequential perception dataset (SPD). The distribution of tracked objects is relatively balanced.}
	\label{fig:distribution for spd}
\end{figure}

\subsection{The Trajectory Forecasting Dataset}\label{sec:tra-pd}
We are also interested in studying trajectory forecasting to predict the future locations of tracked objects. Accurately predicting the behavior of surrounding traffic participants can facilitate more rational decision-making and improve the safety of autonomous driving. However, the ego-vehicle prediction capabilities are significantly limited by the lack of sufficient perceptual information and the lack of interaction between different traffic participants. It is valuable to study the Vehicle-Infrastructure Cooperation (VIC) trajectory forecasting to fully utilize the infrastructure data and improve the forecasting ability.
Although the Sequential Perception Dataset (SPD) can be used to study the VIC trajectory forecasting, the data scale needs to be larger, and the richness of the trajectories needs to be higher to explore various behaviors. Therefore, we mined interesting trajectories from 336 driving and 336 infrastructure hours at 28 urban intersections in the Beijing Yizhuang Area to form a large-scale trajectory dataset. Details about the data collection and trajectory mining are presented in the Appendix.

\begin{figure}[t]
	\centering
    \includegraphics[width=0.45\textwidth]{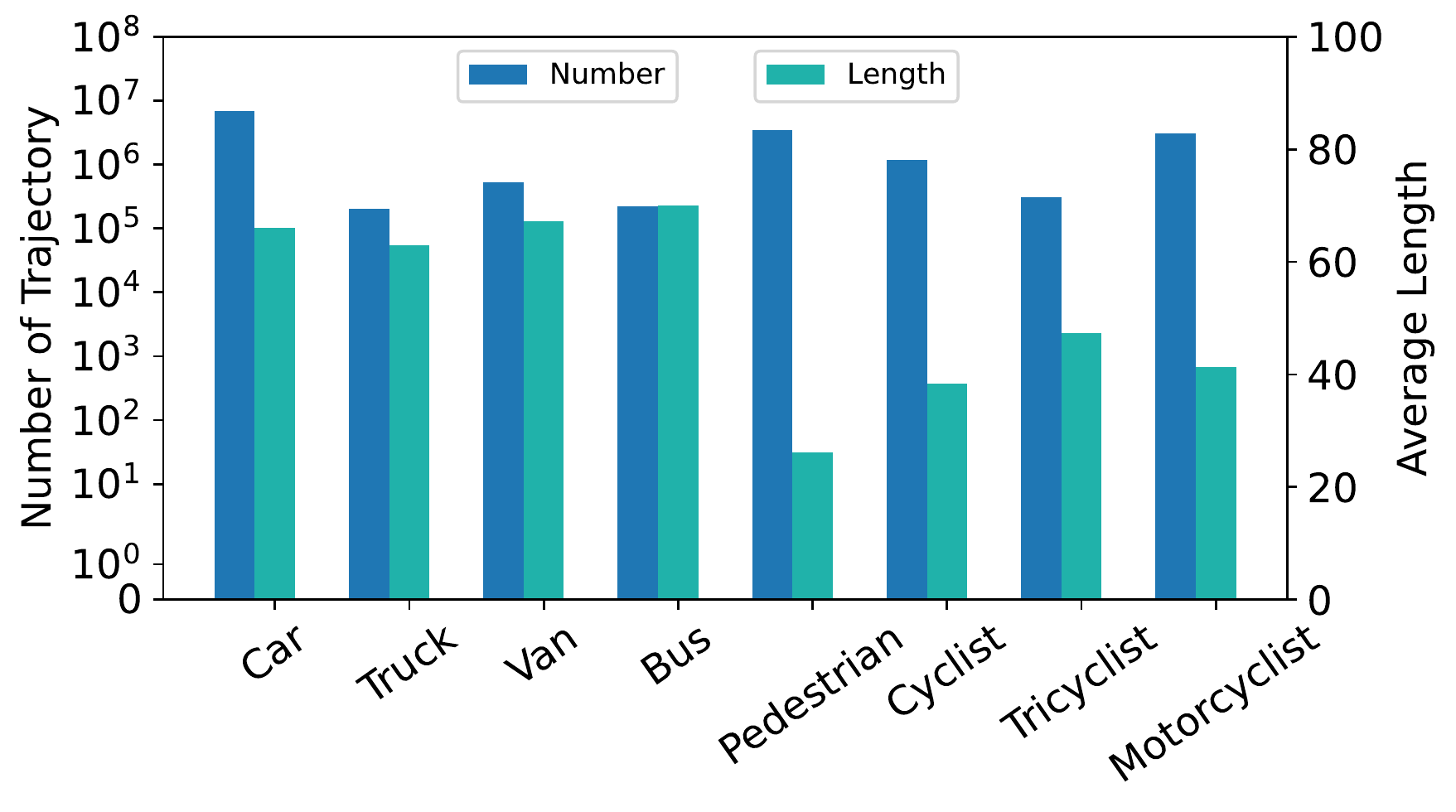}
    \vspace{-15pt}
	\caption{Total number and average length per category for the trajectory forecasting dataset in a relatively uniform distribution of trajectory categories.}
	\label{fig: distribution for sfd}
\end{figure}

The Trajectory Forecasting Dataset (TFD) is composed of about 50,000 cooperative-view, 80,000 infrastructure-view and 80,000 ego-vehicle-view scenarios. Each scenario includes a sequence of tracked object data for 10 seconds at 10HZ, a local vector map, and real-time traffic light signals (only provided for cooperative-view and infrastructure-view scenarios). Among them, 50,000 cooperative-view scenarios were collected at the same time and intersection, where the ego vehicle drove through the equipped intersections.
The tracked objects contain 3D boxes modeled with 7 dimensions, an object type attribute from 8 classes, and a trajectory ID. Additionally, we provide cooperative trajectory annotations for cooperative-view scenarios. The cooperative trajectory is generated in a similar way to cooperative tracking annotation but without manual refinement. Each cooperative trajectory is marked with which trajectories it originated from.
The released dataset is diverse in terms of different classes and locations. The distribution of classes is presented in Figure~\ref{fig: distribution for sfd}.
We provide the detailed data collection and generation in the Appendix.

\subsection{Vector Maps and Traffic Lights.}\label{sec:hd-map}
We provide vector maps for the areas covering the selected 28 intersections, organized similarly to Argoverse~\cite{chang2019argoverse}. The vector maps contain lane centerlines, crosswalks, and stoplines, represented by line segments with starting and ending points. To meet data security requirements, we add a constant offset to the coordinates of the points located in the world coordinate system. For each lane centerline, we provide attributes such as turning left or right, and we also provide the actual lane width so that we can calculate the boundaries of each lane. As traffic vehicles must follow the lane, including the centerline and boundary, to obey traffic rules, building the spatial context between trajectories and vector maps can provide valuable hints for trajectory tracking and forecasting.

Additionally, we provide real-time traffic light signals for the infrastructure portion of the Trajectory Forecasting Dataset (TFD). During the collection and storage of infrastructure sensor data, we also record traffic light data at 10 Hz. The traffic light signals include the timestamp, location, color status, shape status, and time remaining. This information can significantly influence the behavior of traffic participants. It is worth noting that although nuPlan~\cite{caesar2021nuplan} also provides traffic light data, their data is estimated offline based on traffic flow statistics, whereas our data is obtained directly from the traffic lights themselves.

\section{VIC3D Tracking Task}
\begin{figure*}[t]
	\centering
	\includegraphics[width=0.9\textwidth]{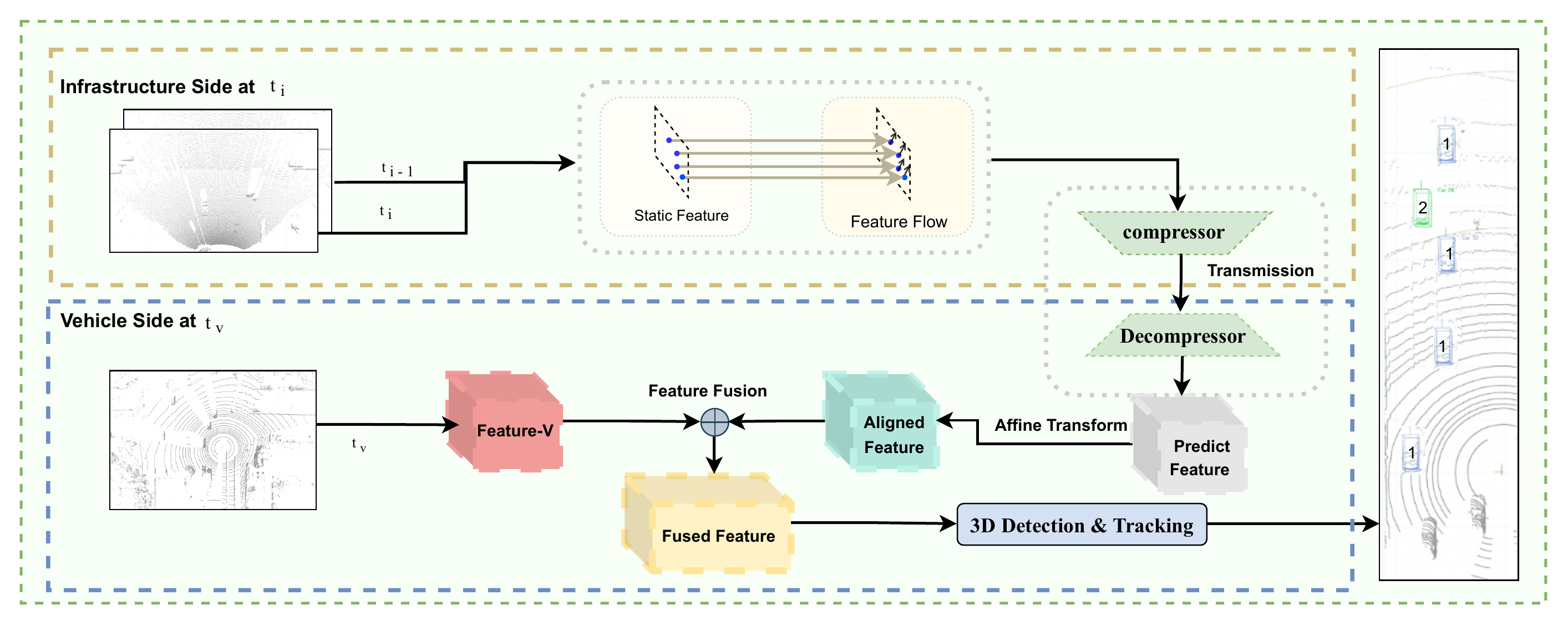}
    \vspace{-12pt}
	\caption{Overview of the FF-Tracking framework. The framework transmits compressed features and feature flows, which can effectively reduce transmission costs while removing fusion errors caused by communication delays.}
	\label{fig:fftracking-framework}
\end{figure*}
In this section, we detail the formalization of the Vehicle-Infrastructure Cooperative 3D (VIC3D) Tracking task, along with the corresponding evaluation metrics. Furthermore, we propose the FF-Tracking framework, which builds upon FFNet~\cite{yu2023vehicle}, to address the issue of degraded tracking performance caused by latency, thereby improving the overall efficiency of VIC3D Tracking.

\paragraph{Task Description.} 
VIC3D Tracking aims to cooperatively locate, identify, and track 3D objects using both infrastructure and ego-vehicle sequential data while operating under limited communication bandwidth. The input for VIC3D Tracking consists of sequential frames from both ego-vehicle and infrastructure sources:
\begin{itemize}
\item Ego-vehicle sequential frames ${I_v(t_v^{'})|t_v^{'}\leq t_v}$ as well as its relative pose ${M_v(t_v^{'})|t_v^{'}\leq t_v}$: captured at and before time $t_{v}$, where $I_{v}(\cdot)$ denotes the capturing function of ego-vehicle sensors.
\item Infrastructure sequential frames ${I_i(t_i^{'})|t_i^{'}\leq t_i}$ as well as its relative pose ${M_i(t_i^{'})|t_i^{'}\leq t_i}$: captured at and before time $t_{i}$, where $I_{i}(\cdot)$ denotes the capturing function of infrastructure sensors.
Here, $t_i$ should be earlier than $t_v$ (i.e., $t_i< t_v$) due to the communication delay.
\end{itemize}
The outputs of VIC3D Tracking include the category, location, orientation, and unique tracking ID of each object in the area of interest surrounding the ego vehicle over time $t_v$. The corresponding ground truth is the set of 3D tracked objects appearing in one of the cooperative-view sensors over time $t_v$, which can be formulated as:
\begin{equation}
GT = (GT_{v} \cup GT_{i}) \cap R,
\end{equation}
where $GT_{v}$ is the ground truth for ego-vehicle sensor perception, $GT_{i}$ is the ground truth for infrastructure sensor perception, and $R$ is the ego-vehicle interest region.

\paragraph{Evaluation Metrics and Analysis.}
VIC3D Tracking has two primary objectives: achieving better tracking performance while minimizing transmission costs to reduce bandwidth consumption. To assess these objectives, we use the following metrics:
\begin{itemize}
\item MOTA, MOTP and IDS: Multi-Object Tracking Accuracy (MOTA), Multi-Object Tracking Precision (MOTP), and ID Switch (IDS) are three commonly used evaluation metrics for 3D tracking \cite{geiger2013vision,cae2020nus}. We use these metrics to measure the performance of VIC3D Tracking approach.
\item BPS: Byte Per Second (BPS) measures the amount of data transmitted from the infrastructure to the ego vehicle per second, taking into account the transmission frequency.
\end{itemize}

However, achieving these objectives presents several challenges. Firstly, we need to reduce the amount of data transmitted to meet the limited communication bandwidth requirement, while ensuring that the transmitted data are valuable enough to improve the tracking performance. The intermediate form is the most likely to achieve a balance between performance and transmission among the three possible data transmission forms (raw, intermediate, and perceived data). Secondly, latency can cause significant damage to cooperative fusion due to scene changes and dynamic object movements over time. Hence, we should consider the use of prediction alignment to remove fusion errors.

\paragraph{FF-Tracking Framework.} 
To address the challenges of VIC3D Tracking, we propose a middle fusion framework called FF-Tracking, which is based on feature flow prediction in FFNet~\cite{yu2023vehicle}. FF-Tracking transmits both feature and feature flow instead of the single static feature from the infrastructure to the ego vehicle.
We predict the future feature to align with the ego-vehicle timestamp using the following linear estimation:
\begin{equation}\label{eq:feature flow}
F_{future}(t) = F_0 + t * F_1,
\end{equation}
where $F_0$ denotes the static feature and $F_1$ denotes the feature flow. With the predicted feature, we can effectively address the fusion error and solve the latency challenges.
To further reduce the transmission cost, we compress the features and feature flows before transmitting them. This approach enables us to achieve the goals of better tracking performance and lower transmission cost while meeting the limited communication bandwidth requirement.

The FF-Tracking framework consists of following parts.
1) Extracting the feature and feature flow from past sequential infrastructure frames.
2) Compressing, transmitting, and decompressing the static feature and feature flow.
3) Predicting the infrastructure feature using Eq.~\ref{eq:feature flow}. 
4) Fusing the features. We transform the predicted feature into a local ego-vehicle coordinate system and then fuse it with the ego-vehicle feature.
We extract the ego-vehicle feature from the ego-vehicle point clouds.
5) Generating the tracking results. We use a Single Shot Detector (SSD)~\cite{Yang2020ssd} to generate the 3D object outputs and then use the AB3DMOT~\cite{weng2020ab3dmot} to track the objects and assign a unique tracking ID for each object.
The whole process is also illustrated in Fig.~\ref{fig:fftracking-framework}.
Please refer~\cite{yu2023vehicle} to more feature flow prediction configurations.

\section{VIC Trajectory Forecasting Tasks}
In this section, we present two trajectory forecasting tasks based on the trajectory forecasting dataset: Online-VIC Forecasting and Offline-VIC Forecasting. These tasks aim to investigate how to effectively leverage real-time infrastructure information and offline behavior knowledge transfer from the infrastructure to the vehicle side.

\subsection{Online-VIC Forecasting Task}
\paragraph{Task Formulation.}
Online-VIC Forecasting can be formulated as the problem of predicting future trajectories using real-time infrastructure and vehicle-side data. The inputs for Online-VIC Forecasting are:
\begin{itemize}
\item A set of infrastructure trajectories $\{T_i^{(l)}(t_i)\}$  and traffic light signals, where the trajectory $T_i^{(l)}(t_i)$ contains the sequential coordinates of agent $A_i^{(l)}$ at and before time $t_i$.
\item Local vector maps.
\item A set of ego-vehicle trajectories $\{T_v^{(k)}(t_v)\}$, where the trajectory $T_v^{(k)}(t_v)$ contains the sequential coordinates of agent $A_v^{(k)}$ at and before time $t_v$. Note that $t_i$ should be earlier than $t_v$ due to the latency. However, in this paper, we ignore the latency to explore how to integrate infrastructure information better and consider $t_i$ equal to $t_v$.
\end{itemize}
The output is the specified target agent's future coordinates for time steps ${t=t_v+1, \cdots, t_{pred}}$.
To make the forecasting task more challenging, we predict longer trajectories and define the forecasting task as observing the past 50 frames (5$s$) and then predicting the future 50 frames (5$s$).

\vspace{-2mm}
\paragraph{Evaluation Metrics and Analysis.}
In autonomous driving, there are often diverse possible future behaviors of traffic participants. Therefore, we output multiple possible future trajectories for each target agent for evaluation. Similar to Argoverse~\cite{chang2019argoverse}, we use the minimum Average Displacement Error (minADE), minimum Final Displacement Error (minFDE), and Missing Rate (MR) as the metrics to measure the prediction performance. We evaluate the model with Top-$K$ predictions as our metrics, where $K=6$.

Our approach is based on the Trajectory Forecasting Dataset (TFD), which involves receiving and fusing infrastructure data from an intersection environment with complicated traffic situations. There are several challenges to achieving better prediction performance. One of these challenges is to effectively utilize valuable infrastructure information to enhance the incomplete perception results of the vehicle side, which is limited due to the single-vehicle view. Another challenge to establish a proper social context by incorporating infrastructure-perceived agents for better reasoning about the future behaviors of the target agent. Finally, it is crucial to improve the encoding of vector maps and traffic light signals to better assist in prediction.

\subsection{Offline-VIC Forecasting Task}
The Offline-VIC Forecasting task aims to transfer knowledge extracted from various infrastructure sequences to predict ego-vehicle trajectories. During inference, the model can only utilize the ego-vehicle data and cannot access real-time infrastructure data, similar to the traditional trajectory forecasting task~\cite{chang2019argoverse}. Similar to Online-VIC Forecasting, we define the prediction task as observing the past 50 frames (5 seconds) and predicting the future 50 frames (5 seconds). We measure the prediction results using minADE, minFDE, and MR metrics and evaluate the model with Top-$K$ predictions, where $K$=6. The main challenge in solving this task is extracting appropriate knowledge from heterogeneous infrastructure data for transfer.

\section{Experiments}
\subsection{VIC3D Tracking Benchmarks}
\begin{figure}[t]
	\centering
	\includegraphics[width=0.4\textwidth]{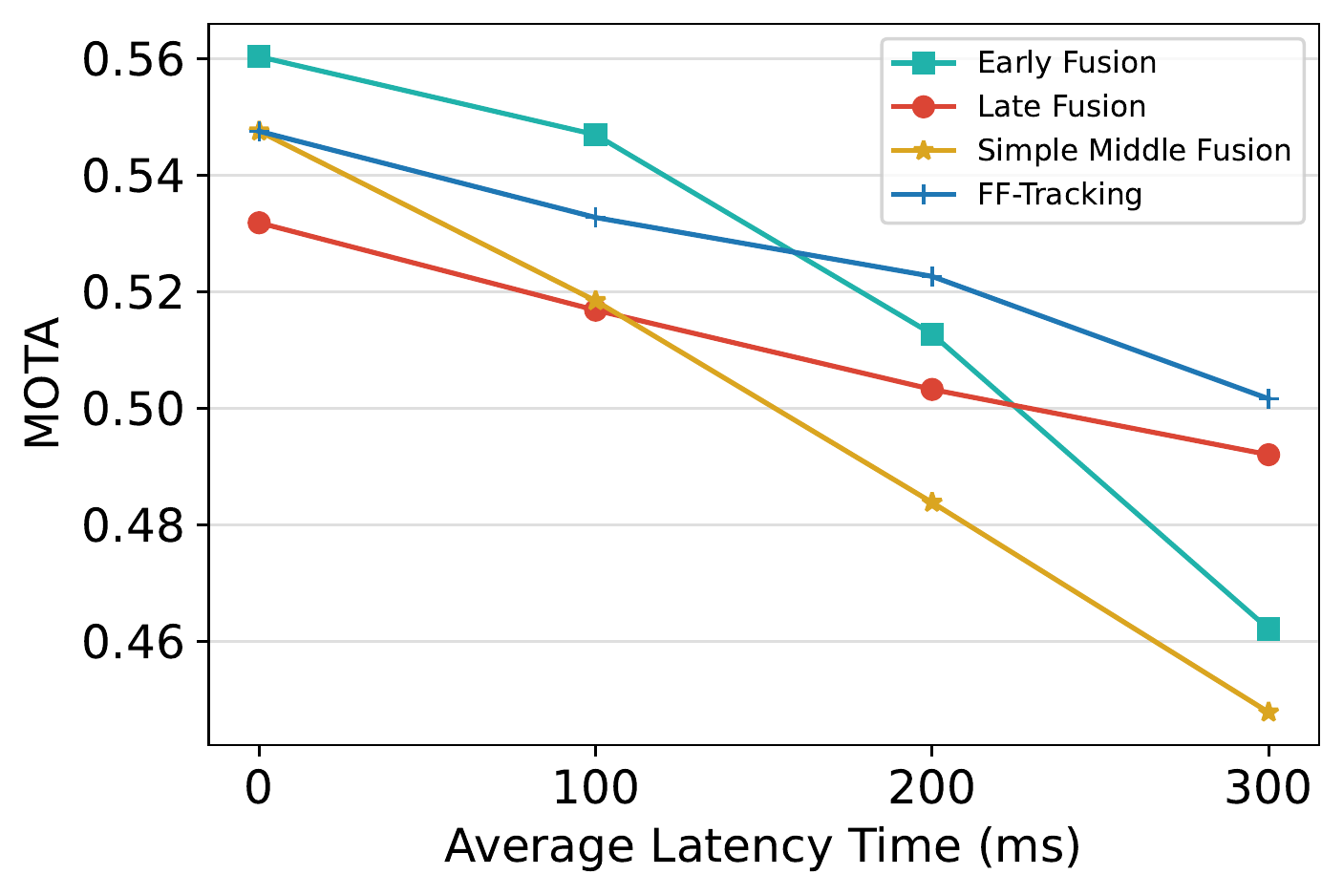}
    \vspace{-10pt}
	\caption{Comparison of VIC3D Tracking Baseline Models with Varying Latencies. Our proposed FF-Tracking model demonstrates greater robustness to latency when compared to the early fusion, late fusion, and simple middle fusion models.}
	\label{fig: exp-vic3d-tracking-latency}
\end{figure}

In this section, we present the results of our extensive experiments, which include different fusion approaches, input modalities, and latency settings. The experiments are conducted on the sequential perception dataset (SPD), and the train/valid/test split ratio is set to 5:2:3.
We only consider four classes of Car, Van, Bus and Truck and the objects located in a rectangular of [0, -39.68, 100, 39.68].
The results are summarized in Table~\ref{tab: VIC3D Tracking Benchmark results} and visualized in Figure~\ref{fig: exp-vic3d-tracking-latency}.

\begin{table*}[htpb!]
\centering
\small
\caption{\textbf{Evaluation Results for VIC3D Tracking on SPD at Different Latency Levels.} The "Vehicle Only" approach utilizes only ego-vehicle data, while "Concat fusion" combines pseudo images generated from point clouds. The evaluation of tracking performance employs three metrics: MOTA, MOTP, and IDS. Additionally, the transmission cost per second is assessed using the BPS metric. Notably, \textbf{in this experiment we only compare MOTA scores for the evaluation} and do not consider the MOTP and IDS scores for comparison.}
\renewcommand\arraystretch{1.25}
\label{tab: VIC3D Tracking Benchmark results}
\scalebox{0.90}{
\begin{tabular}{c|ccc|ccc|c}
\hline
\hline
Modality&  Latency ($ms$) &  Fusion Type & Fusion Method & MOTA~$\uparrow$~\text{     } & MOTP~\text{     }  & IDS~\text{     } & BPS (Byte/s)~$\downarrow$\\
\hline
\hline
\multirow{2}*{Image}& 0 & Vehicle Only &- &10.96 &58.69  &2 &0\\
 & 0 & Late Fusion & Hungarian~\cite{Kuhn2010TheHM}  &22.27&57.25&194 &3.3$\times 10^3$ \\
 \hline
 \hline
\multirow{4}*{PointCloud} & 0 &Vehicle Only & - &39.31 &67.28&109 & 0 \\
 & 0 &Early Fusion &  Concat & \textbf{56.03} &70.17&296 &1.3$\times 10^7$\\
 & 0 &Late Fusion & Hungarian~\cite{Kuhn2010TheHM} &53.18 &72.35&273&3.3$\times 10^3$\\
 &0 &  Middle Fusion & V2VNet~\cite{Wang2020V2VNetVC} &54.75&69.76&222  &6.2$\times 10^5$\\
 \hline
\hline
\multirow{1}*{PointCloud}
 & 0 & Middle Fusion & \textbf{FF-Tracking}  & 54.75 & 69.76 & 222 & 6.2$\times 10^5$\\
\hline
\hline 
\multirow{3}*{PointCloud} 
 & 200 & Early Fusion & Concat  &51.27 &69.67& 234 & 1.3$\times 10^7$\\
 & 200 &Late Fusion & Hungarian~\cite{Kuhn2010TheHM} & 50.32 &71.58& 260 &3.3$\times 10^3$\\
 & 200 &Middle Fusion &  V2VNet~\cite{Wang2020V2VNetVC} &48.38 & 68.99 &231  &6.2$\times 10^5$\\
\hline
\hline
\multirow{1}*{PointCloud}
 &200 & Middle Fusion & \textbf{FF-Tracking}  & \textbf{52.26} &69.64&225 &1.2$\times 10^6$\\
\hline
\hline
\end{tabular}
}
\end{table*}

\subsubsection{Baselines}
The VIC3D Tracking problem can be tackled using three solutions: early fusion, middle fusion, and late fusion. Early fusion involves fusing infrastructure raw data, middle fusion fuses intermediate-level infrastructure data like feature maps, and late fusion fuses infrastructure perception results.
Raw data contains all information but requires the highest transmission cost, while perception results consume the least amount of transmission cost but lose valuable information. We conducted experiments to evaluate the performance of these fusion solutions for VIC3D Tracking.

\paragraph{Solution with Middle Fusion.}
We implemented the FF-Tracking model and a simple middle fusion model to explore middle fusion with intermediate data. We first explain how to train the FF-Tracking model. 
We pre-trained the FF-Tracking model on the training part of the sequential perception dataset for 40 epochs without considering latency. The learning rate was set to 0.001, and the weight decay was set to 0.01. We fine-tuned the FF-Tracking model on the training part of thesequential perception dataset for 20 epochs by adding random latency. The learning rate was set to 0.001, and the weight decay was set to 0.01. 
We then applied V2VNet~\cite{Wang2020V2VNetVC} as a simple middle fusion model to solve VIC3D Tracking and compared it with FF-Tracking. Compared to FF-Tracking, the V2VNet~\cite{Wang2020V2VNetVC} only transmits a single feature and keeps the other configurations the same as FF-Tracking. We trained the model for 40 epochs with a learning rate of 0.001 and a weight decay of 0.01.
Note that FF-Tracking incurs a higher transmission cost per second compared to simple middle fusion due to the requirement of transmitting additional feature flow. Furthermore, in 0$ms$ latency, FF-Tracking degenerates into V2VNet, which suggests that FF-Tracking and V2VNet manifest equivalent tracking performance under 0$ms$ latency conditions.

\paragraph{Solution with Early Fusion.}
We implement early fusion with point cloud inputs. First, we convert the infrastructure point cloud into the ego-vehicle coordinate system. Then, we convert both infrastructure and ego-vehicle point clouds into pseudo-images and fused them. We used PointPillars~\cite{lang2019pointpillars} as a detector to generate 3D outputs and AB3DMOT~\cite{weng2020ab3dmot} to track each object. We directly train and evaluate the detector with the fused point cloud. Additionally, we also evaluate the model with different latencies.

\paragraph{Solution with Late Fusion.}
To investigate the fusion effect with perception results, we implement late fusion using point cloud and image inputs. Specifically, we employ PointPillars~\cite{lang2019pointpillars} to locate and identify objects from both infrastructure sequential frames and ego-vehicle sequential frames. Additionally, we use ImvoxelNet~\cite{rukhovich2021imvoxelnet} to perceive 2D objects from the infrastructure and ego-vehicle sequential images.
Next, we transmit the infrastructure objects to the ego vehicle and fuse them with the ego-vehicle objects based on Euclidean distance measurements. Then we use AB3DMOT~\cite{weng2020ab3dmot} to track the fused objects. Finally, we evaluate the model's performance with different latencies.

\subsubsection{Analysis}
\paragraph{V2X view vs. Single-vehicle view.}
In Table~\ref{tab: VIC3D Tracking Benchmark results}, we present the evaluation results for both fusion and no-fusion methods. When using point clouds as input, all fusion methods outperform the no-fusion strategy, even when there is a performance drop due to communication delay. For instance, with point cloud as input and 200$ms$ latency, the early fusion method improves the MOTA (multiple object tracking accuracy) of vehicles by 11.96\% (from 39.31\% to 51.27\%). 
Thus, vehicle-infrastructure cooperative perception can effectively enhance 3D tracking performance.

\paragraph{Middle Fusion vs. Early Fusion\&Late Fusion.}
We compared the performance of middle fusion, early fusion, and late fusion techniques using point cloud as input and with 0$ms$ latency. Our results indicate that early fusion achieves higher tracking performance than middle fusion (56.03\% vs. 54.75\% MOTA), while middle fusion requires less transmission cost (6.2$\times 10^5$ Byte/s and 1.2$\times 10^6$ Byte/s vs. 1.3$\times 10^7$ Byte/s). Although late fusion requires the least transmission cost with 3.3$\times 10^3$ Byte/s, it still achieves lower tracking performance than middle fusion (53.18\% vs. 54.75\% MOTA). Our findings suggest that the middle fusion technique can achieve a better balance between transmission cost and tracking performance.

\paragraph{FF-Tracking can overcome the latency challenge.}
We present evaluation results of different fusion methods with a 200$ms$ latency, as shown in Table~\ref{tab: VIC3D Tracking Benchmark results}. All the fusion methods show a performance drop as the latency increases. For example, the early fusion has a 4.76\% MOTA drop, and the simple middle fusion has a 6.37\% MOTA drop when the latency is increased from 0$ms$ to 200$ms$. In comparison, our FF-Tracking model only has a 2.49\% MOTA drop.
We also present additional evaluation results at different latencies in Fig.~\ref{fig: exp-vic3d-tracking-latency}. Our FF-Tracking model remains robust to all latencies, and importantly, outperforms early fusion by up to 4\% MOTA at 300$ms$ latency. Additionally, our FF-Tracking model achieves the best tracking performance when the latency reaches 200$ms$.

\subsection{Trajectory Forecasting Benchmarks}
This section provides the baselines for solving the Online-VIC and Offline-VIC forecasting tasks on the trajectory forecasting dataset (TFD) with a train/val/test split of 5:2:3. The evaluation results are presented in Table~\ref{tab: Online Experiment 1}.

\subsubsection{Baselines}
We choose TNT~\cite{zhao2021tnt} and HiVT~\cite{zhou2022hivt} as base models and train them with different configurations. We encode only the trajectories and vector maps that are within 50m of the ego vehicle. We evaluate the models on val part of 50,000 cooperative-view dataset. Specifically:
\begin{itemize}
\item \textbf{Baseline 1:} We only use ego-vehicle data and vector maps from the 50,000 cooperative-view data. We train the TNT~\cite{zhao2021tnt} and HiVT~\cite{zhou2022hivt} models for 30 epochs, and the other settings remain the same as the original.
\item \textbf{Baseline 2:} We use vector maps and both ego-vehicle and infrastructure trajectories. We propose the PP-VIC framework, a simple yet effective hierarchical perception-prediction method for solving the Online-VIC Forecasting task. Firstly, we first use CBMOT~\cite{benbarka2021score} to fuse the infrastructure and ego-vehicle trajectories. We only fuse or add infrastructure trajectories that are relatively complete or have very high detection scores. Then, we apply TNT~\cite{zhao2021tnt} and HiVT~\cite{zhou2022hivt} to encode the trajectories and vector maps to generate future trajectories, respectively. We train the PP-VIC model for 30 epochs,  and the other settings remain the same as the original.
\item \textbf{Baseline 3:} We use ego-vehicle data and vector maps from the 50,000 cooperative-view data and additionally use 80,000 infrastructure-view trajectories. We pre-train the TNT~\cite{zhao2021tnt} on the 80,000 infrastructure trajectories and then fine-tune the TNT~\cite{zhao2021tnt} initialized with the pre-trained models. We train HiVT~\cite{zhou2022hivt} in the same way.
\end{itemize}

\begin{table}[htpb!]
\centering
\footnotesize
\caption{Evaluation results for different baselines. Using infrastructure trajectories can improve forecasting performance.}
\renewcommand\arraystretch{1.3}
\label{tab: Online Experiment 1}
\scalebox{0.9}{
\begin{tabular}{cc|ccc}
\hline
\hline
Using Infrastructure   & Prediction & \multicolumn{3}{c}{K = 6} \\
\cline{3-5}
Trajectories &  Model & minADE~$\downarrow$ & minFDE~$\downarrow$ & MR~$\downarrow$ \\
\hline
\hline
 \XSolid  &  TNT~\cite{zhao2021tnt} & 12.01 & 24.15 & 0.84 \\
 Online  & TNT~\cite{zhao2021tnt} & 8.27 & 17.25 & 0.76 \\ 
   Offline & TNT~\cite{zhao2021tnt} & 4.36 & 9.23 & 0.62 \\ 
\hline
\hline
\XSolid  &  HiVT~\cite{zhou2022hivt} & 1.55 & 2.59 & 0.36 \\
  Online  & HiVT~\cite{zhou2022hivt} & 1.27 & 2.36 & 0.35 \\ 
  Offline & HiVT~\cite{zhou2022hivt} & 1.52 & 2.27 & 0.30 \\ 
\hline
\hline
\end{tabular}
}
\end{table}

\subsubsection{Analysis}
\paragraph{Online infrastructure trajectories are useful.} 
Compared to baselines that do not use any infrastructure information, PP-VIC achieves lower minADE, minPDE, and MR. PP-VIC with TNT~\cite{zhao2021tnt} achieves a minADE that is 3.74 lower than the TNT~\cite{zhao2021tnt} model that does not use infrastructure trajectory information, and PP-VIC with ~\cite{zhou2022hivt} achieves a minADE that is 0.28 lower than the ~\cite{zhou2022hivt} model that does not use trajectory information. These results suggest that online utilization of infrastructure trajectories can improve forecasting performance.

\paragraph{Offline infrastructure trajectories are useful.} 
TNT~\cite{zhao2021tnt} pretrained on extra infrastructure trajectories achieves a 7.65 minADE reduction compared to TNT~\cite{zhao2021tnt} without the use of any infrastructure data.
 HiVT~\cite{zhou2022hivt} pretrained on extra infrastructure trajectories achieves a 0.03 minADE reduction compared to HiVT~\cite{zhou2022hivt} without the use of any infrastructure data.
 The experimental results demonstrate that extracting knowledge from infrastructure trajectories can effectively improve forecasting performance.

\section{Conclusion}
This paper presents a large-scale sequential V2X dataset, where all the data elements, including data frames, trajectories, vector maps, and traffic lights, are captured and generated from natural scenery. The paper introduces three new tasks for the vehicle-infrastructure cooperative autonomous driving community to better study how to utilize infrastructure information to improve sequential perception and trajectory forecasting ability. Several benchmarks are carefully designed for the fair evaluation of the introduced tasks.
The experimental results demonstrate that infrastructure data can improve tracking and trajectory forecasting ability. Moreover, this paper proposes a novel FF-Tracking approach to solve the VIC3D Tracking problem.

\section*{Acknowledgements} 
This work was supported by Baidu Inc. through the Apollo-AIR Joint Research Center, and partially supported by the General Research Fund of HK under Grants No. 27208720 and No. 17200622. The authors would like to express their gratitude to the Beijing High-level Autonomous Driving Demonstration Area and Beijing Academy of Artificial Intelligence for their support throughout the dataset construction and release process.

\newpage
\balance
{\small
\bibliographystyle{ieee_fullname}
\bibliography{egbib}
}

\end{document}


\title{Appendix for V2X-Seq Dataset and Benchmarks}

\maketitle

\begin{abstract}
In this appendix, we provide further details on the V2X-Seq dataset. Specifically, Section~\ref{sec: v2x-seq-dataset-deployment} outlines the sensor deployment and intersection layout in detail. Section~\ref{sec: v2x-seq-dataset-trajectory-mining} describes the trajectory collection and mining process. Finally, in Section~\ref{sec: v2x-seq-dataset}, we have included selected examples from the released dataset along with their visualizations.
\end{abstract}

\section{Sensor Deployment and Intersection Layout}\label{sec: v2x-seq-dataset-deployment}
\begin{figure}[htbp]
	\centering
	\includegraphics[width=0.5\textwidth]{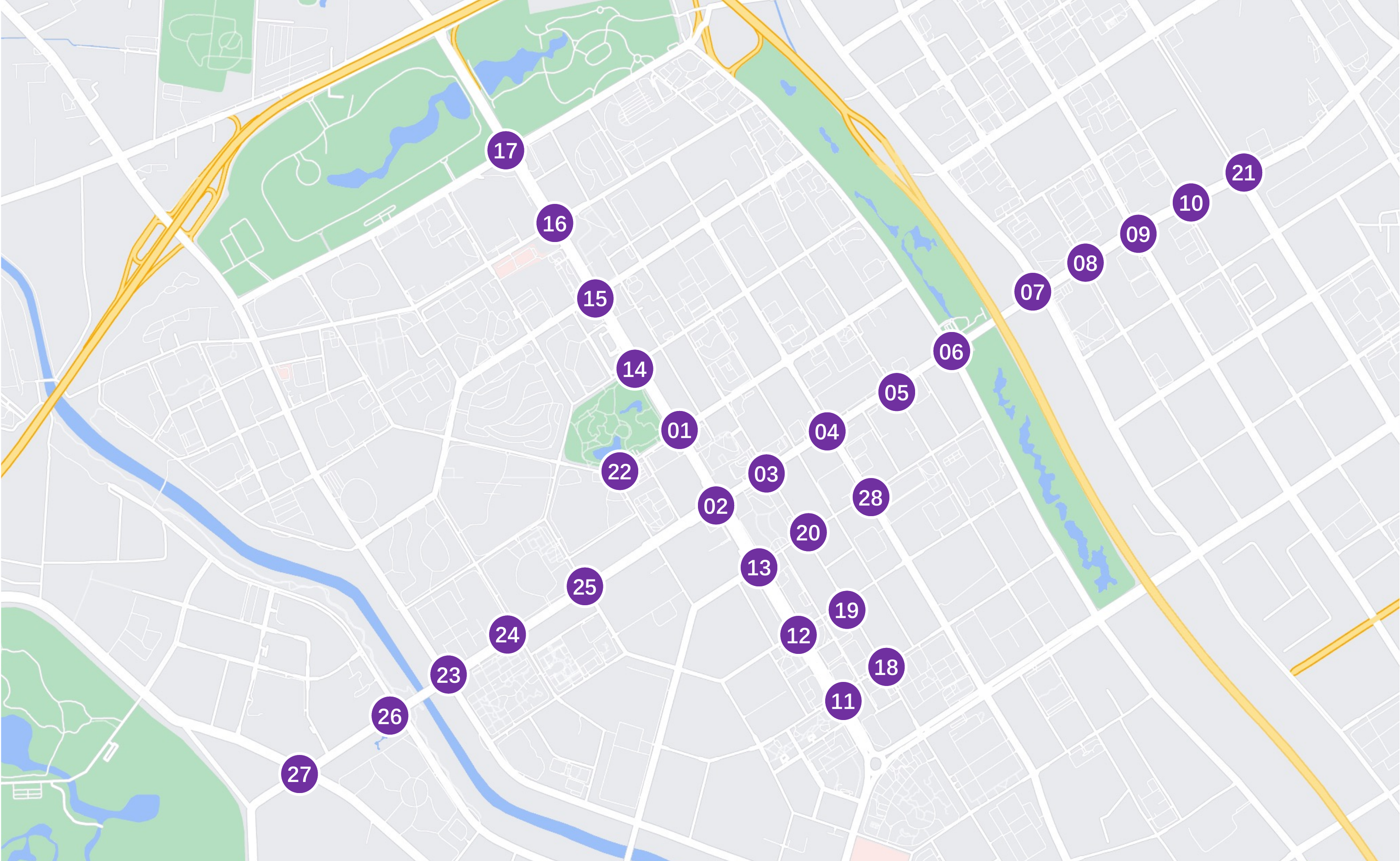}
	\caption{Layout of 28 Deployed Urban Intersections.}
     \label{fig: intersection layout}
\end{figure}

We selected 28 urban traffic intersections in Beijing and deployed sensors at these locations. The layout of these selected intersections can be seen in Figure~\ref{fig: intersection layout}.

\begin{figure}[htbp]
	\centering
	\includegraphics[width=0.4\textwidth]{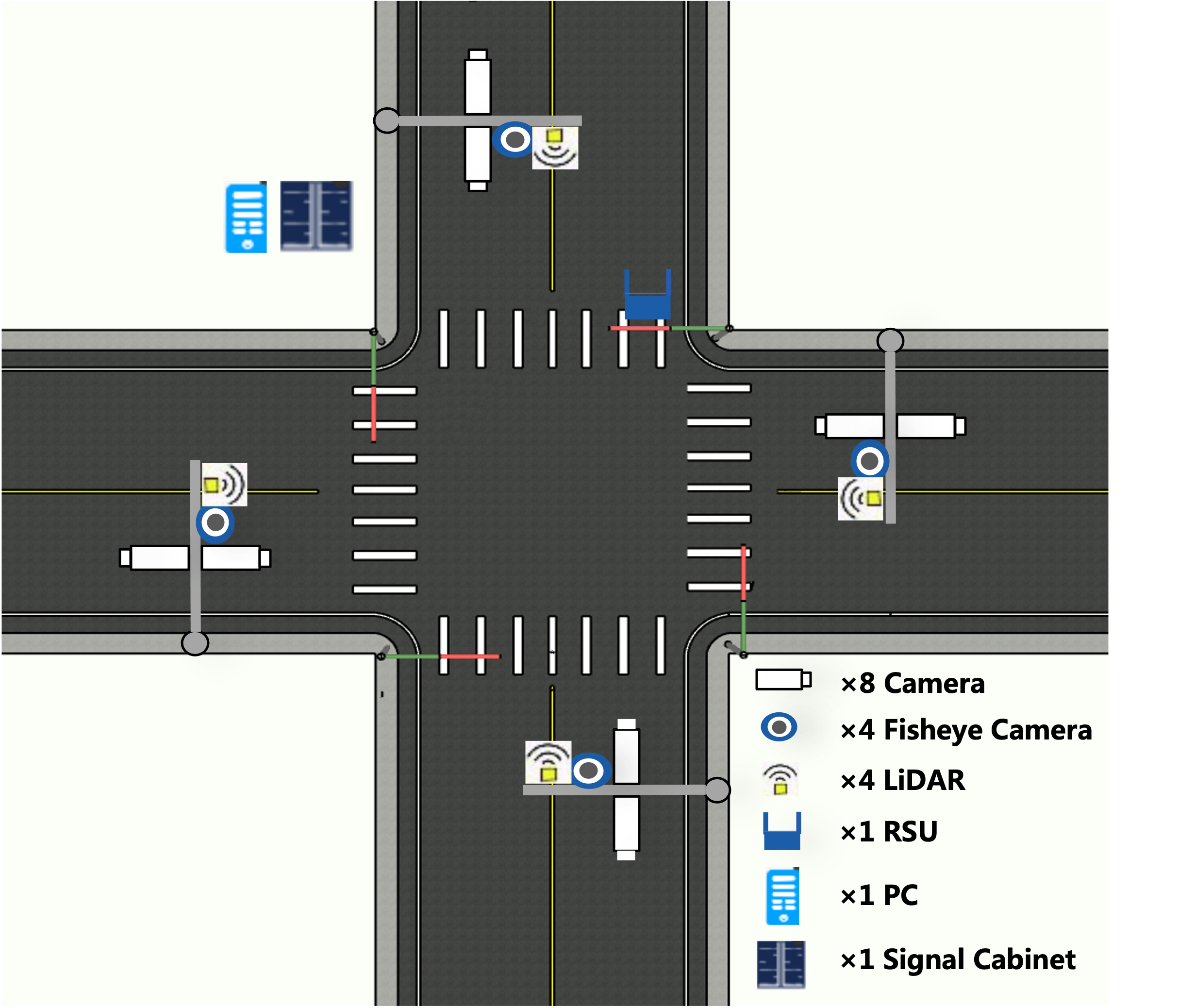}
	\caption{The Infrastructure Sensor Deployment.}
	\label{fig: infrastructure sensor deployment}
\end{figure}

\begin{figure}[htbp]
	\centering
	\includegraphics[width=0.4\textwidth]{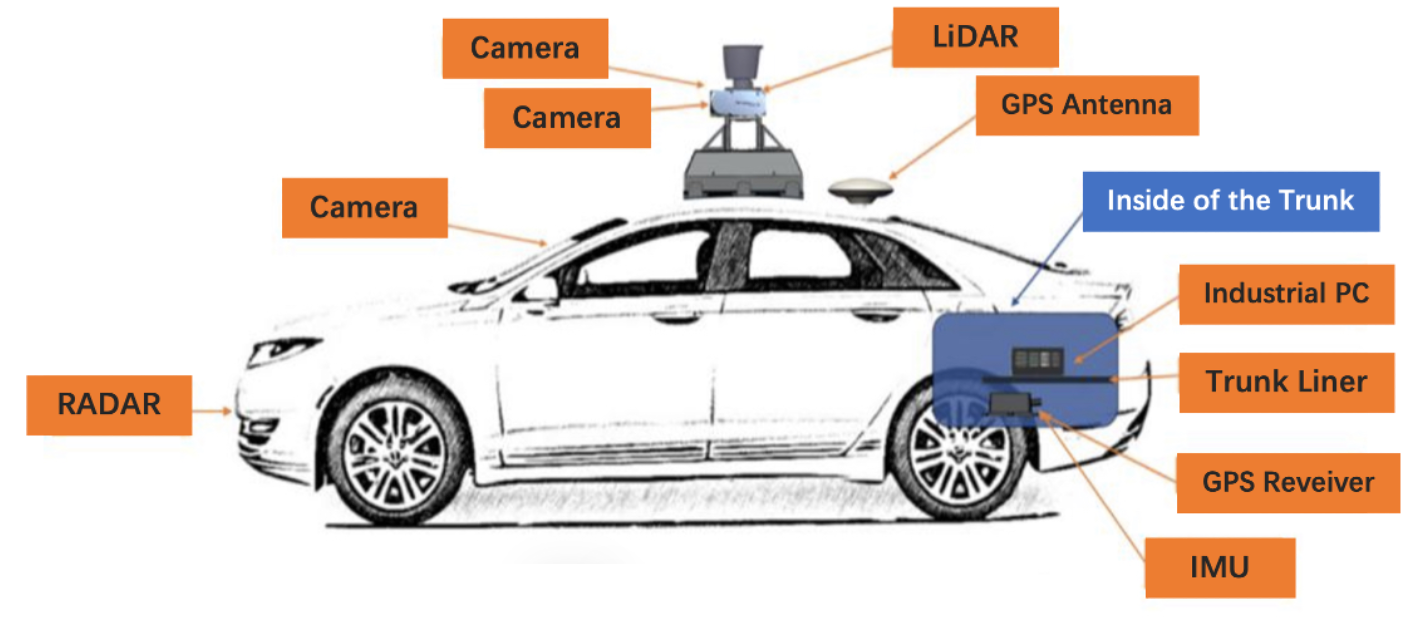}
	\caption{The Vehicle Sensor Deployment.}
	\label{fig: vehicle sensor deployment}
\end{figure}

We deploy 4$\sim$6 pairs of 300-beam LiDAR and high-resolution cameras for each intersection. These infrastructure sensors can fully cover the intersection areas.
We provide the configuration of infrastructure sensor deployment in Fig.~\ref{fig: infrastructure sensor deployment}.
We deploy one 40-beam LiDAR and six high-quanlity cameras for the self-driving vehicle. We provide the vehicle sensors deployment in Fig.~\ref{fig: vehicle sensor deployment}.

\section{Trajectory Collecting and Mining}\label{sec: v2x-seq-dataset-trajectory-mining}
In this section, we explain the process of building the trajectory forecasting dataset, with a particular focus on the 50,000 cooperative-view scenarios.

To collect the sensor data, we drove our self-driving vehicle through sensor-equipped areas, collecting both infrastructure-side and vehicle-side sensor data over a period of 672 hours. This data was saved every three minutes.
We then only input this infrastructure images and vehicle sensor data into trained 3D object detection and tracking models to generate trajectory sequences consisting of 3D boxes, each with a class attribute from 8 categories and a unique trajectory ID. These boxes were uniformly transformed from local coordinates into world coordinates, resulting in the trajectory sequences repository.
Finally, we mined interesting trajectory segments from the repository to create about 50,000 cooperative-view scenarios.

The trajectory mining process consisted of several steps, including scene fragmentation, scene selection, trajectory fusion, trajectory scoring, filtering, and attribute creation.
First, we fragmented the infrastructure-side and vehicle-side sequences into 10-second segments, with 5 seconds of overlap between adjacent segments.
Next, we selected the infrastructure-side and vehicle-side segments that corresponded to the same equipped intersection to form segment pairs.
We then fused the infrastructure-side and vehicle-side trajectories, generating cooperative-view trajectories for each segment pair. We followed the method of generating cooperative annotation for VIC3D tracking, but filtered out any matching with low scores and directly discarded them. Each cooperative trajectory was connected to the origin-view trajectory IDs.
Next, we assigned a score to each cooperative trajectory based on various factors such as turning, speeding up, slowing down, lane changing, and completion.
Finally, we kept 50,000 sequences with high-score trajectories. We set one to five trajectory in each segment as the target agent type, and these trajectories were located within a certain range of the ego vehicle. Other trajectories were set as other agent types.
Additionally, we mined more trajectories from the trajectory repository to increase the size of the infrastructure-side and vehicle-side sequences to about 80,000 each.

\section{Visual Example for V2X-Seq}\label{sec: v2x-seq-dataset}
\begin{figure}
	\centering
	\includegraphics[width=0.5\textwidth]{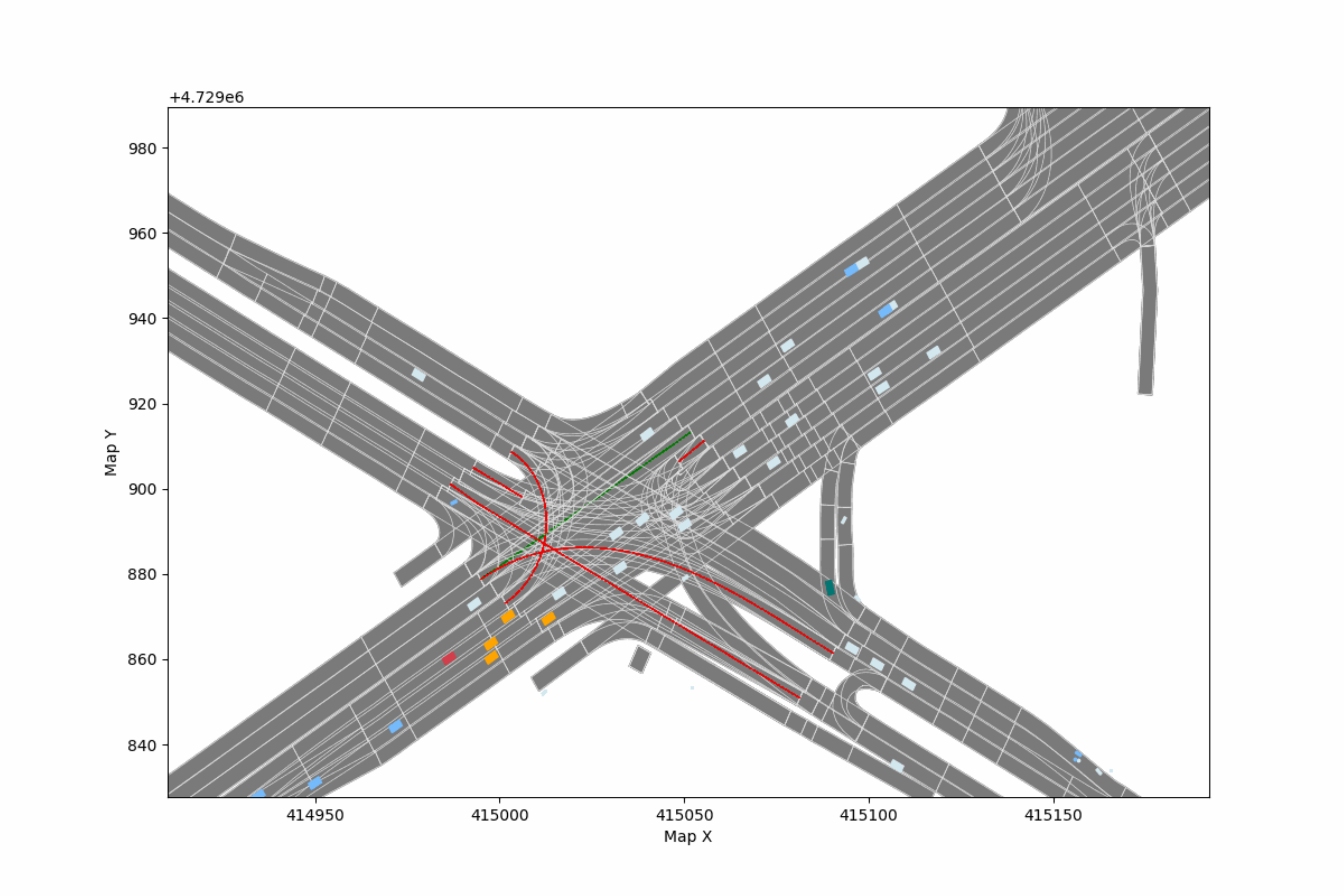}
	\caption{ Screenshot of an interesting scenario from the Trajectory Forecasting Dataset with Cooperative-view. The green point denotes the ego vehicle. The red point denotes the first target agent, while the orange points denote the second to fifth target agents. The light blue boxes represent the objects generated from the vehicle side, while the dark blue boxes represent the complementary objects with infrastructure data. The red lines indicate the red traffic light for the located lanes.}
     \label{fig:vis-for-tfd}
\end{figure}

\begin{figure*}[t]
    \centering
    \begin{subfigure}[a]{0.69\textwidth}
           \centering
           \includegraphics[width=\textwidth]{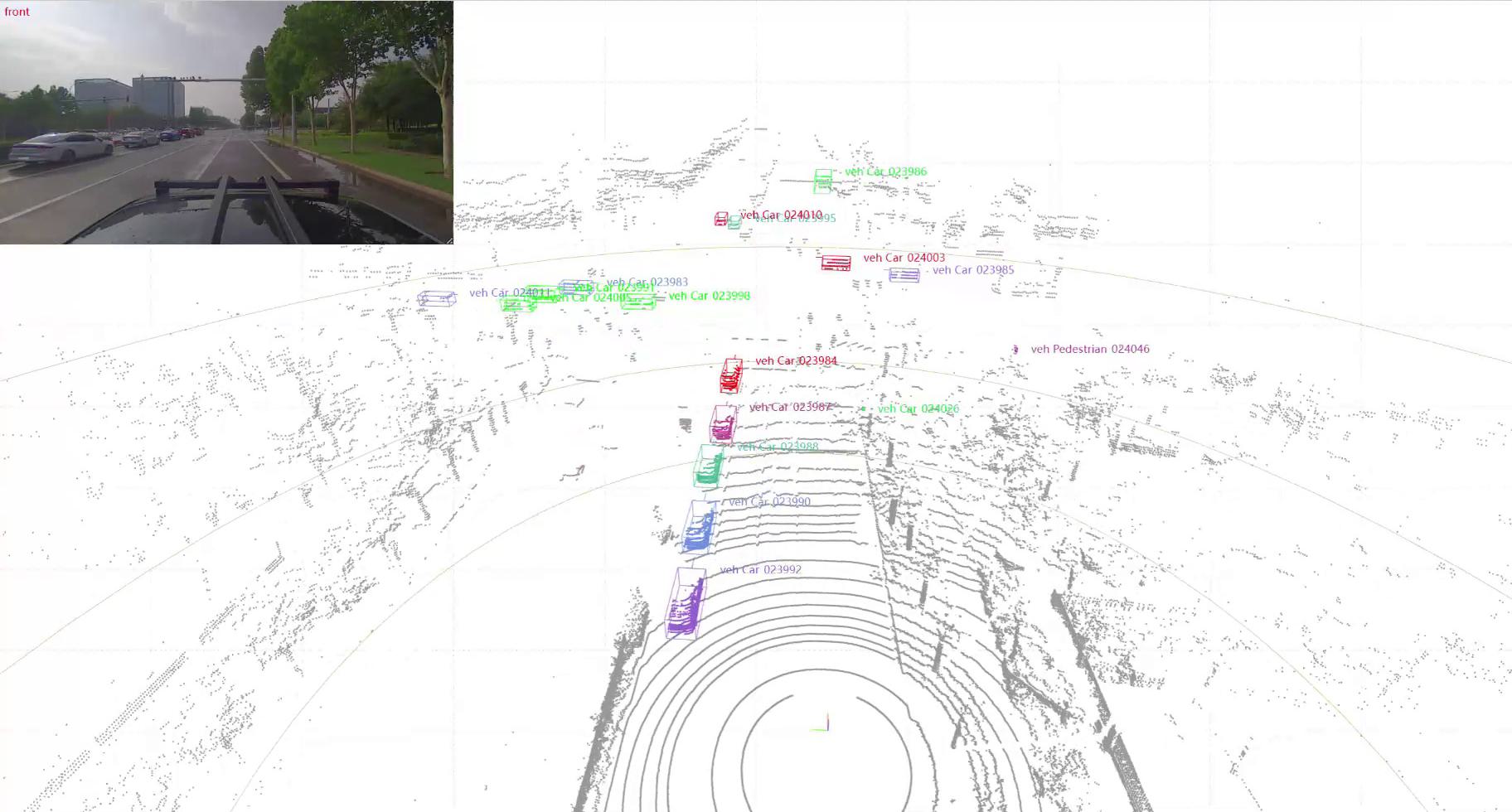}
            \caption{Vehicle-view Visualization.}
    \end{subfigure}
    
    \begin{subfigure}[b]{0.69\textwidth}
            \centering
            \includegraphics[width=\textwidth]{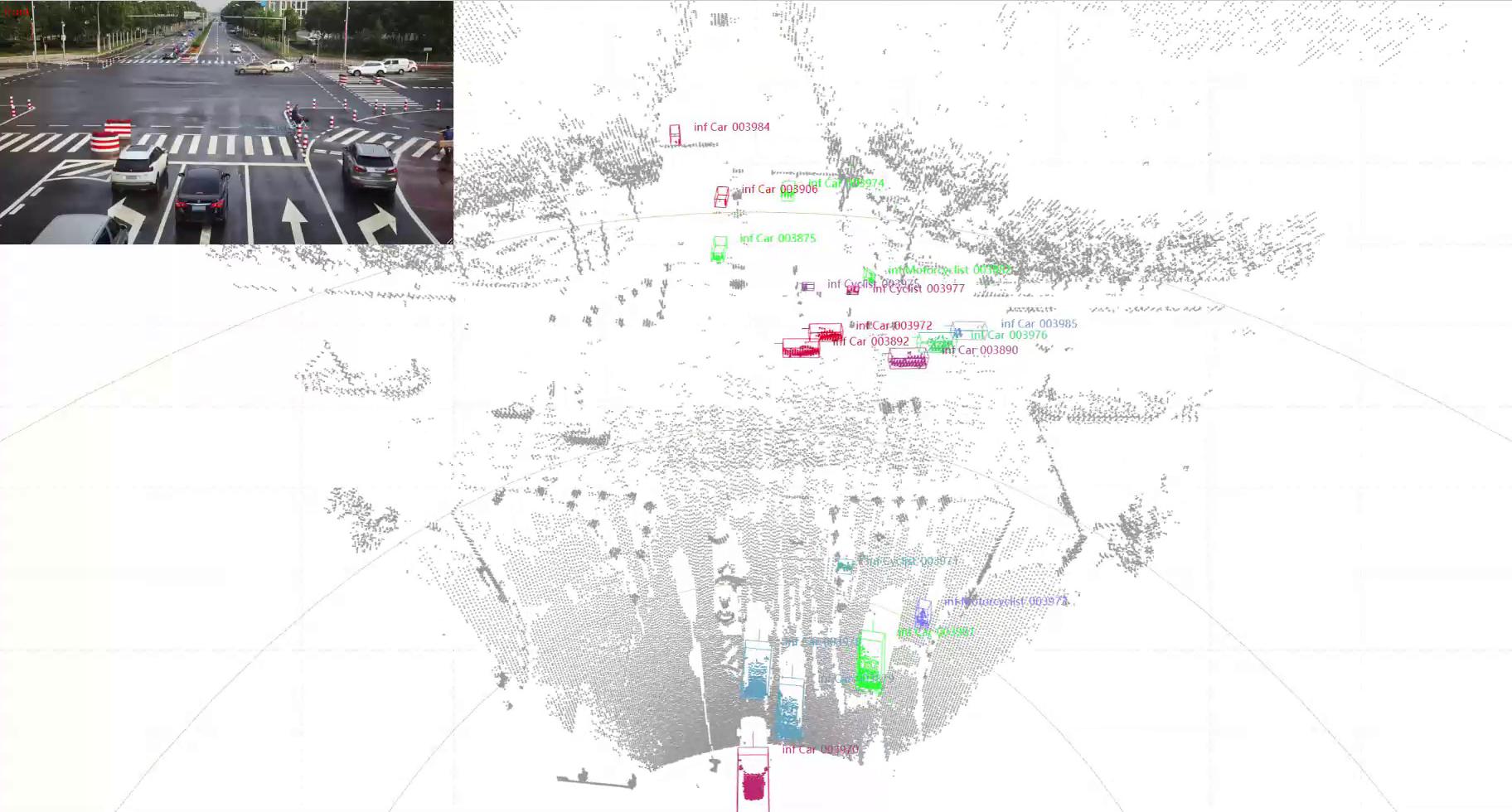}
            \caption{Infrastructure-view Visualization.}
    \end{subfigure}

    \begin{subfigure}[c]{0.69\textwidth}
            \centering
            \includegraphics[width=\textwidth]{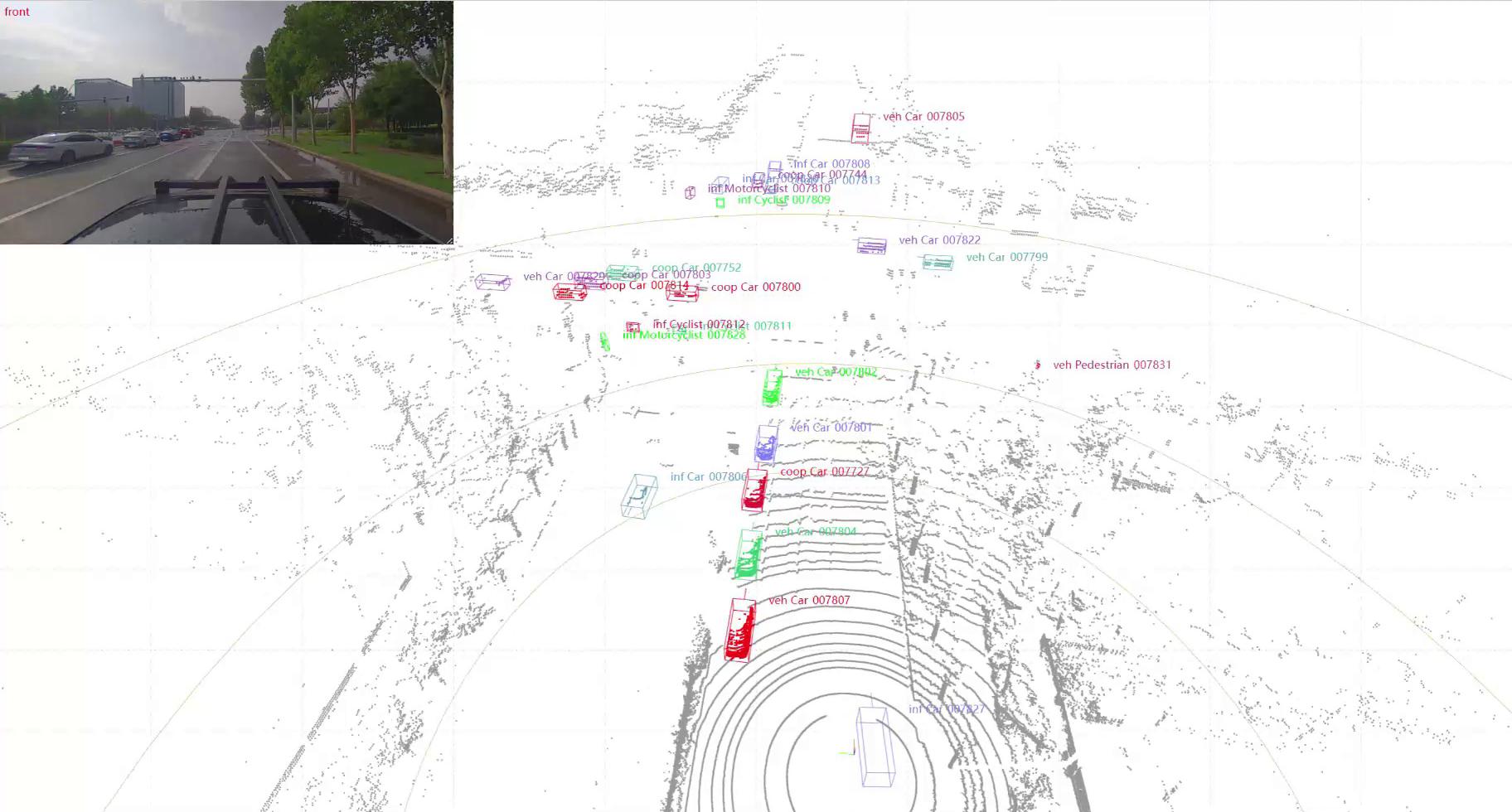}
            \caption{Cooperative-View Visualization. We visualize the cooperative annotation on vehicle-side images and point clouds. Each 3D box is marked with a label indicating whether it is from the infrastructure-side or vehicle-side annotations. The cooperative view allows us to see more objects and obtain a more comprehensive understanding of the scene.}
    \end{subfigure}
    
    \caption{Visualization Example of the Sequential Perception Dataset. Each 3D box is marked with its attribute and tracking ID.}\label{fig:visual-example-spd}
\end{figure*}

We present a interesting scenario that were mined from the trajectory forecasting dataset in Figure~\ref{fig:vis-for-tfd}. Additionally, we provide a visualization example for the sequential perception dataset in Figure~\ref{fig:visual-example-spd}.

